\documentclass[final]{cvpr}

\usepackage{times}
\usepackage{epsfig}
\usepackage{graphicx}
\usepackage{amsmath}
\usepackage{amssymb}
\usepackage{algorithm}
\usepackage{algorithmic}
\usepackage{tabularx}

\def\xx{\mathbf{x}}
\def\wn{\mathbf{w}}
\def\yy{\mathbf{y}}
\def\cc{c^{'}}
\def\cM{c^{''}}
\def\ff{f^{'}}
\def\hh{h^{'}}
\def\ww{w^{'}}

\def\R{{\rm I\!R}}
\def\chm{\checkmark}
\def\tclr{\textcolor}

\usepackage[pagebackref=true,breaklinks=true,colorlinks,bookmarks=false]{hyperref}

\begin{document}

\title{TRAT: Tracking by Attention Using Spatio-Temporal Features}

\author{\parbox{16cm}{\centering
    {\large Hasan Saribas$^1$, \hspace{0.2cm} Hakan Cevikalp$^2$, \hspace{0.2cm} Okan K\"op\"ukl\"u$^3$, \hspace{0.2cm} Bedirhan Uzun$^2$}\\
    {\normalsize
    \vspace{0.2cm}
    $^1$ Eskisehir Technical University \\
    $^2$ Eskisehir Osmangazi University \\
    $^3$ Technical University of Munich}}
}

\maketitle

\begin{abstract}
Robust object tracking requires knowledge of tracked objects' appearance, motion and their evolution over time. Although motion provides distinctive and complementary information especially for fast moving objects, most of the recent tracking architectures primarily focus on the objects’ appearance information. In this paper, we propose a two-stream deep neural network tracker that uses both spatial and temporal features. Our architecture is developed over ATOM tracker and contains two backbones: \mbox{(i) 2D-CNN} network to capture appearance features and \mbox{(ii) 3D-CNN} network to capture motion features. The features returned by the two networks are then fused with attention based Feature Aggregation Module (FAM). Since the whole architecture is unified, it can be trained end-to-end. The experimental results show that the proposed tracker \textbf{TRAT} (TRacking by ATtention) achieves state-of-the-art performance on most of the benchmarks and it significantly outperforms the baseline ATOM tracker. The source code and models are available at \small \url{https://github.com/Hasan4825/TRAT}.

\end{abstract}

\section{Introduction}
The visual object tracking is an important computer vision task, and the goal is to track a target object in the subsequent frames of a video where the target object is identified in the first frame. It is widely used in many domains including surveillance, video and activity analysis, and robotics. However, it is a challenging task since there are numerous factors such as significant deformation and appearance variations of the target object, illumination changes, background clutter, occlusion, etc. that can make the tracking of the target extremely difficult. The main difficulty arises from the fact that the tracker must learn an appearance model of the target object at the initial frame just by using bounding box information. Then, it must have the adaptability to generalize to all variations of the target object appearances in the subsequent frames. 
\begin{figure}[h!]
		\begin{center}
				\includegraphics[width=\columnwidth]{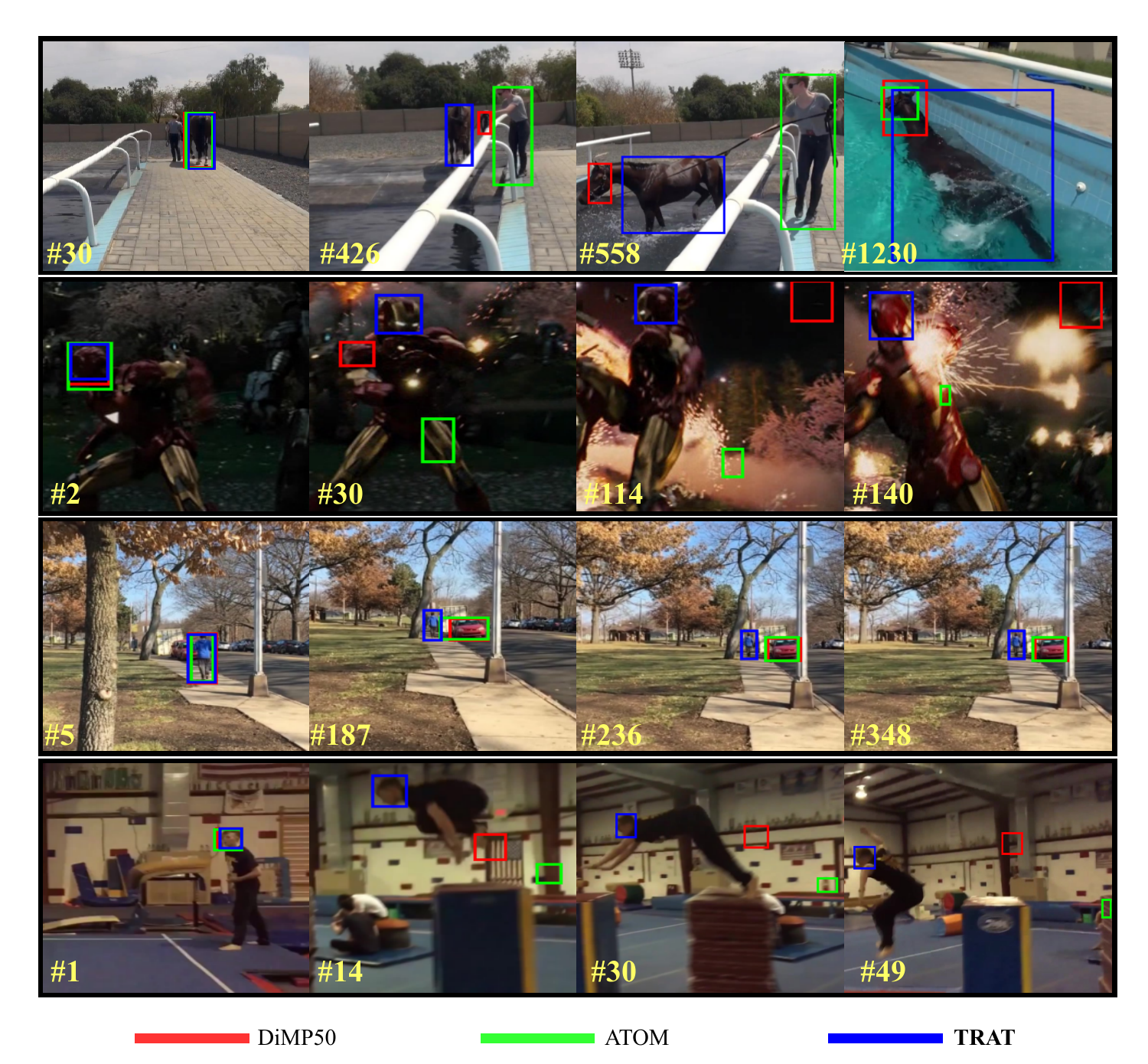}
				\caption{Visual comparison of the TRAT with state-of-the-art trackers. ATOM \cite{R97} and DiMP50 \cite{R100} trackers are based on only the appearance information coming from 2D-CNN features. Consequently, they fail to track fast moving objects. Our approach TRAT employs 3D-CNN and 2D-CNN backbones to simultaneously extract appearance and motion information and provide accurate bounding box predictions for slow and fast moving objects.}
					\label{fig:VisualComparison}
		\end{center}		
\end{figure}

Object tracking requires the incorporation of spatial and temporal features captured within video frames. The spatial part carries information about the scenes and object appearances in individual video frames, whereas the temporal part contains the motion information of the tracked objects across the video frames. So far, the majority of the trackers that employ deep neural network architectures only used spatial features coming from video frames for tracking. On the other hand, spatio-temporal features have been widely used for both activity recognition and detection in video frames \cite{R62,R63,R64,R66,R67}. However, capturing spatio-temporal features is not very straightforward in tracking settings as in activity recognition since we do not have access to all video frames during online tracking. Instead, we have only access to the previous frames of the current frame, and many tracking videos do not include any motion for long time periods. As a result, special attention must be given to combine spatial and temporal information during online tracking. In fact, temporal information is mostly useful when the object appearance changes significantly between consecutive frames because of object speed, shape deformation, illumination variations, and out-of-plane rotations. Motion information also helps when there are similar objects in the vicinity of the tracked object. Fig. \ref{fig:VisualComparison} illustrates several examples where the motion information is critical.

In this paper we propose a novel two-stream deep neural network tracker that uses both spatial and temporal features. We use an attention based Feature Aggregation Module (FAM) to fuse the information coming from two-channels effectively. Proposed FAM makes use of local channel correlations for efficiency and a more discriminative pooling strategy to improve the tracking performance.


\subsection{Related Work}
The most recent state-of-the-art tracking methods can be roughly divided into three categories: CNN (Convolutional Neural Network) based trackers, correlation filter based trackers and Siamese network based trackers. CNN based trackers \cite{R17,R18,R19,R20,R21,R22,R23} usually use shallow neural networks since it has been shown in many studies that the last layers of large CNNs are more effective in capturing semantics. 
Small networks are also necessary for speed issues since updating parameters of large networks is slow and it cannot be accomplished in real-time for online visual tracking. 
There are also methods using larger networks, but they usually update either the last classifier stage or a few last layers of the network during online learning due to the computational efficiency. 
One of the earliest deep neural network learning tracker \cite{R22} used a stacked de-noising auto-encoder to learn generic image features from a large dataset as auxiliary data and then transferred the learned features to the online tracking task. Hong et al. \cite{R23} used CNNs and SVM classifier to learn target-specific saliency maps. Li et al. \cite{R20} introduced an online learning method based on a CNN using multiple image cues. 
Wang et al. \cite{R19} pre-trained a large CNN offline and transferred the learned features to the online tracking as in \cite{R22}. 
Nam and Han \cite{R17} introduced the Multi-Domain Network (MDNet) for object tracking. Beside some online training tricks such as long-term and short-term update strategies, the main novelty of the paper was to show how to transfer rich and effective features for tracking. A fast version of MDNet tracker, called Real-Time MDNet, \cite{R98} used Fast R-CNN \cite{R99} approach to accelerate the slow feature extraction stage of the MDNet method. Cevikalp et al. \cite{R103} proposed a deep neural network tracker using ranking loss which enforces the network to return better bounding boxes framing the target object. For the same purpose, both \cite{R97,R100} used novel deep neural tracking architectures that utilize IoU-Net \cite{R105} whose goal is to estimate and increase the Intersection over Union (IoU) overlap between the target an estimated bounding box to improve the accuracy. 

Correlation filter (CF) based trackers learn a correlation filter to localize the target object in consecutive frames by solving a ridge regression problem efficiently in the Fourier frequency domain. The learned filter is applied to a region of interest in the next frame, and the maximum correlation filter response determines the object location in the new frame. Then, the filter is updated by using this new object location. CF based trackers are extremely fast compared to the deep neural network based trackers owing to the fact that the problem is solved efficiently in the frequency domain. Bolme et al. \cite{R24} introduced a very fast CF based method using the minimum output of squared error (MOSSE) for visual tracking. Kernelized correlation filters using circulant matrices and multi-channel features have been proposed in \cite{R25}. Danelljan et al. \cite{R26} introduced a formulation using continuous convolutional operators, which paved the way for efficiently integrating multi-resolution deep feature maps into the convolution filter based trackers. In \cite{R27}, this method has been improved by the introduction of factorized convolutional operators. Initially, CF trackers used gray levels or hand-crafted features such as histogram of oriented gradients (HOGs) \cite{R25}, color names \cite{R28} or color histograms \cite{R35}. On the other hand, CF trackers proposed in \cite{R18,R32} utilized CNN features extracted from pre-trained CNNs. The most recent studies introduced methods to learn both deep CNN features and correlation filters simultaneously \cite{R29,R30,R31}. The best performing tracker \cite{R31} of the VOT2017 \cite{R34} challenge uses this methodology. 
ROI Pooled Correlation Filter (RPCF) method \cite{R81} used a region of interest (ROI) based pooling operation in the correlation filters algorithm. Wang et al. \cite{R82} integrated forward tracking and backward verification steps, which are based on forward-backward trajectory analysis, into the unsupervised training phase. Xu et al. \cite{R83} proposed a feature compression in both spatial and channel dimensions by using group channel feature selection method to learn correlation filters. \cite{R84} used adaptive spatially regularized correlation filters model to estimate the object location and scale. Saribas et al. \cite{R85} and Huang et al. \cite{R86} achieved high accuracy and speed on tracking of objects in aerial videos by using correlation filters.

The trackers that are based on Siamese networks on the other hand are built based on distance (or similarity) metric learning for visual object tracking \cite{R36,R37,R38,R39}. Initially, these methods \cite{R36,R37} are trained on larger datasets such as ILSVRC15 \cite{R124} and the learned metric (i.e., matching function) is simply evaluated online during tracking, and these methods did not have the ability to update the previously learned distance function during tracking. Later, Guo et al. \cite{R39} and Wang et al. \cite{R38} proposed methods that update the learned distance function online during tracking.
One of the main challenges in Siamese-based trackers is how to adapt aspect ratio changes. To address this problem, Siamese Region Proposal Network (SiamRPN) \cite{R87} used region proposal networks. SiamRPN++ \cite{R88} proposed a layer-wise feature aggregation structure to improve performance and reduce the model size. To solve the data imbalance problem of RPN, Fan and Ling \cite{R89} have used hard negative sampling process and cascaded RPN instead of one-stage RPN. Gradient-Guided network (GradNet) method \cite{R90} used a novel gradient-guided network to avoid overfitting and increase adaptation ability. SiamMask \cite{R91} utilized a multi task learning approach to solve both object tracking and object segmentation simultaneously. Wang et al. \cite{R93} used a two-stage series-parallel matching for tracking where the coarse matching (CM) and fine matching (FM) stages enhance the robustness performance and discrimination power of Siamese networks. Choi et al. \cite{R95} presented a novel meta-learner network by adding target aware feature space to improve tracking accuracy.

\smallskip\noindent{\bf Trackers Using Spatio-Temporal Features:} Regarding trackers using both spatial and temporal features, \cite{R68} showed that a correlation filter tracker using deep motion (optical flow) features combined with deep RGB features outperforms the same tracker using appearance (deep RGB features) information alone. Teng et al. \cite{R70} proposed a deep neural network that includes temporal and spatial networks, where the temporal network collects key historical temporal samples by solving a sparse optimization problem. The output of the temporal network is fed to the spatial network that further refines the localization of tracked target. A Graph Convolutional Tracking (GCT) method built based on Siamese framework, which uses spatio-temporal features, has been given in \cite{R71}. Zhang et al. \cite{R72} learned a spatio-temporal context model between the tracked target and its local surrounding background and used it for robust tracking of the targets. The method which is the most related to our proposed method is the FlowTrack introduced in \cite{R69}. FlowTrack also uses both spatial and temporal information for better tracking of the object. 
However, there are significant differences between the FlowTrack and our tracker TRAT. FlowTrack is a CF based tracker  by making use of several 2D-CNNs to extract RGB and optical flow modality features, and an attention module to fuse them. 
In contrast, we use a 3D-CNN backbone to extract temporal features in addition to the 2D-CNN backbone which extract spatial features. The feature maps coming from these two backbones are fused with an effective and discriminative attention based feature aggregation module using local channel correlations. The reason to include 3D-CNN backbone in our tracker is that 3D convolutional kernels inherently captures pixel movements (i.e., motion information) on consecutive frames. Although 3D-CNNs are considered computationally heavy, they are much more efficient compared to optical flow calculation. To the best of our knowledge, TRAT is the first architecture, which employs 3D-CNNs to capture motion information for the task of object tracking.

\begin{figure}[tb]
		\begin{center}
				\includegraphics[width=1.22\columnwidth]{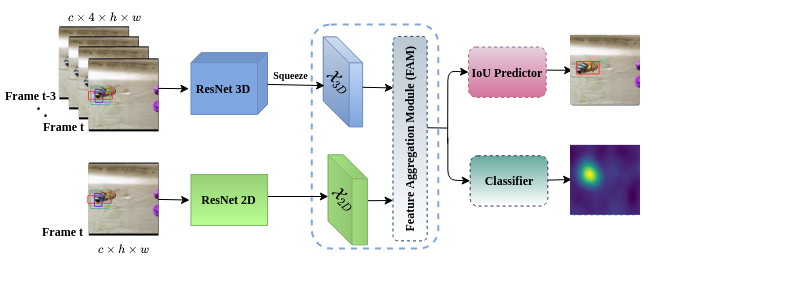}
				\caption{The architecture of the proposed two-stream tracker. It uses both spatial and temporal features through 2D-CNN and 3D-CNNs. The feature maps of the two-networks are fused by using a feature aggregation module shown with dashed line rectangle. }
				\label{fig:tracker}
				\vspace{-0.2cm}
		\end{center}		
\end{figure}

\section{Method}
We propose a two-stream deep neural network tracker for visual object tracking, which is illustrated in Fig.~\ref{fig:tracker}. Our tracker TRAT is developed over ATOM tracker architecture in a way that it uses a two-stream network including both 3D-CNN and 2D-CNN architectures rather than using 2D-CNN alone. In addition, we propose an efficient feature aggregation module (FAM) to aggregate the feature maps of 3D-CNN and 2D-CNN backbones to track the target object more successfully. Therefore, the proposed tracker mainly includes 3 important components as seen in the figure: 3D-CNN, 2D-CNN, and feature aggregation module (FAM). In the following, we will first briefly describe ATOM (Accurate Tracking by Overlap Maximization) tracker and explain each  proposed components in details. Next, we give the implementation details at the end of this section.

\subsection{ATOM}
ATOM tracker includes two components designed exclusively for target estimation and classification. Target estimation module is learned offline and it predicts the Intersection over Union (IoU) ratio between the target and estimated bounding box. Target classification module uses a discriminative correlation filter (DCF) to return an initial bounding box candidate which will be used in target estimation module. Then, 10 object bounding box proposals are generated by adding uniform random noise to this initial position. The target estimation module estimates the IoU ratios of these boxes and the final prediction is obtained by taking the mean of the 3 bounding boxes with the highest IoU ratios.

Target classification module of the ATOM consists of a 2-layer fully convolutional neural network defined as,
\begin{equation}
f(\xx;\wn)=\Phi_2(\wn_2*\Phi_1(\wn_1*\xx)),
\end{equation}
where $\xx$ is the backbone feature map, $\wn=\left\{\wn_1,\:\wn_2\right\}$ are the parameters of the network, $\Phi_1, \Phi_2$ are activation functions. The module uses the following discriminative correlation filter objective function given as,
\begin{equation} \label{eq:2}
L(\wn) = \sum_{j=1}^m \gamma_j\left\|f(\xx_j;\wn)-\yy_j)\right\|^2+\sum_k \lambda_k\left\|\wn_k\right\|^2.
\end{equation}
Here, $\yy_j$ represents the classification confidences for feature maps $\xx_j$. The impact of training samples is adjusted by using the weight $\gamma_j$, and $\lambda_k$ is the weight for regularization.

Instead of using only spatial features $\xx$ coming from 2D-CNN as in ATOM, TRAT makes use of spatio-temporal features extracted by 3D-CNN and 2D-CNN, which are fused by attention based FAM.

\subsection{3D-CNN}
Tracking methods using spatial (appearance) information alone struggle in situations where the object appearance changes significantly between consecutive frames because of shape deformation, out-of-plane rotations, illumination variations, and background distractors with similar appearances to the tracked object. In such cases, motion information provides rich complementary information and it may help to disambiguate the target object from the background. Most studies used optical-flow features for motion information. However, in addition to the computational complexity spent for extracting optical-flow features, many tracking videos lack motion for long time periods. As a result, optical-flow features will be completely zero for many video sequences which limits its use for tracking. On the other hand, 3D-CNNs can capture motion information by applying convolution operation in both space and time dimensions. Moreover,3D-CNNs can also provide information when the object is stationary. Therefore, 3D-CNN backbone is selected to be used as motion capturing unit in TRAT architecture. 

In our 3D-CNN, we use 3D-ResNet architecture. During tracking, we give a sequence of cropped ROIs (Region of Interests) to the network as input. The ROIs are cropped from a sequence of consecutive frames in time order and the size of ROIs is determined at the beginning by using the given target ground-truth location in the first frame. Given a cropped ROIs with size $h\times w$, the size of the 3D-CNN input is $\left(c \times f \times h\times w\right)$, where $c=3$, $f$ is the number of consecutive frames, which is set to 4 in our experiments. After convolution operations, 3D-CNN outputs a feature map  $\mathcal{X}_{3D} \in \R^{\cc\times \ff\times \hh\times \ww}$, where $\cc$ is the number of output channels, $\ff=1$, $\hh=\frac{h}{d}$, and $\ww=\frac{w}{d}$. $d$ refers to the applied downsampling rate. The depth dimension of the resulting feature map is reduced to 1 such that output volume is squeezed to  $\left( \cc \times \hh\times \ww\right)$ in order to match the output feature map of 2D-CNN.

\subsection{2D-CNN}
To exploit spatial (appearance) information for tracking, we use a 2D-CNN, which employs ResNet \cite{R106} architecture. In contrast to 3D-CNN, 2D-CNN takes a single ROI image cropped from the most recent frame, and its size is also $h\times w$. The 2D-CNN outputs a feature map $\mathcal{X}_{2D}\in\R^{\cM\times \hh\times \ww}$, where $\cM$ is the number of output channels, and $\hh=\frac{h}{d}$, $\ww=\frac{w}{d}$ as in 3D-CNN output. Since, $\hh$ and $\ww$ are common for both 3D-CNN and 2D-CNNs, the feature maps of these two networks can be concatenated directly. The resulting stacked features include both appearance and motion information and they are passed to the subsequent FAM as input.

\begin{figure*}[tb]
		\begin{center}
				\includegraphics[width=\textwidth]{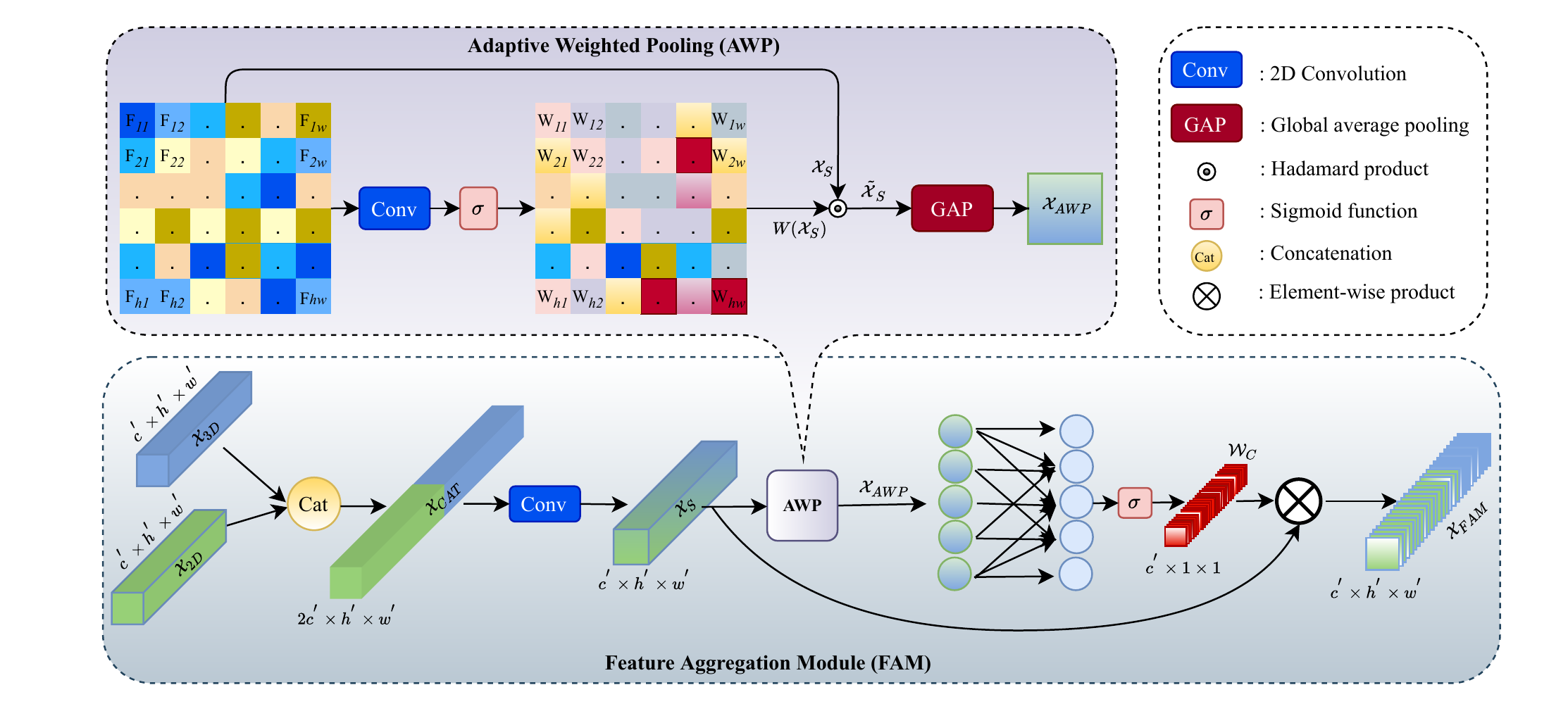}
				\caption{The architecture of the proposed feature aggregation module. }
					\label{fig:attention}
		\end{center}		
\end{figure*}

\subsection{Feature Aggregation Module (FAM)}
The fusion of feature maps coming from 2D-CNN and 3D-CNN backbones should be handled properly in order to achieve superior performance compared to single backbones alone. To this end, we use an effective channel fusion and attention mechanism to aggregate the feature maps. The proposed module is a variant of the Efficient Channel Attention (ECA) module \cite{R109} which can be simply implemented via 1D convolution. As opposed to other channel attention modules that use correlations among all channels, ECA captures only local cross-channel interactions by considering local neighborhoods of each channel. This makes the method both efficient and effective since ECA can be implemented by fast 1D convolutions of size $k$, where kernel size $k$ represents the coverage of local cross-channel interactions.

However, ECA loses discriminative details since it employs global average pooling (GAP) at the beginning, which represents an entire channel with the average sum of the values in the channel. Still, not all values in a feature channel contribute equally to the localization of the tracked object and some features are more discriminative than the others. Therefore, we replace the GAP in ECA module with a more discriminative pooling strategy that uses weighted sum of the values in each channel, which is called as adaptive weighted pooling (AWP) in \cite{R110,R111}. AWP learns to give higher weights to more discriminative features during pooling and this largely increases the detection performance as reported in \cite{R111}. This is somewhat similar to spatial attention since more discriminative features are more heavily weighted during summation.

The proposed attention based feature aggregation module is given in Fig.~\ref{fig:attention}. We use the same architectures for both 3D-CNN and 2D-CNN, therefore the outputs of these networks have same channel size $(\cc=\cM)$. As a result, $\mathcal{X}_{2D}\in\R^{\cc\times \hh\times \ww}$ and $\mathcal{X}_{3D} \in\R^{\cc\times \hh\times \ww}$. 
To fuse feature maps, we utilized  two fusion methods where the first one simply concatenates feature channels whereas the second one adds the feature channels as described below.

\noindent{\bf{Concatenation fusion.}} As given in Fig.~\ref{fig:attention}, our concatenation fusion strategy contains concatenation operation and convolution  operation with $1\times1$ kernel size. The concatenation operation stacks two feature maps along channels to produce $\mathcal{X}_{Cat} = f^{cat}(\mathcal{X}_{2D},\mathcal{X}_{3D}) \in\R^{2\cc\times \hh\times \ww}$. Then, we feed the stacked feature maps into 2D convolutional layer to return to original size by reducing the channel size by half $\mathcal{X}_{S} \in\R^{\cc\times \hh\times \ww}$.

\noindent{\bf{Sum fusion.}} The sum operation simply computes the sum of the two feature maps at the same spatial locations to produce $\mathcal{X}_{S} = f^{sum}(\mathcal{X}_{2D},\mathcal{X}_{3D}) \in\R^{\cc\times \hh\times \ww}$.

After fusion operation, we employ adaptive weighted pooling (AWP) to downscale feature map, $\mathcal{X}_{S}$. To this end, we first feed $\mathcal{X}_{S}$ into convolution layer and apply amplified sigmoid function to the convolved feature map for computing $W(\mathcal{X}_{S})$. In order to obtain new weighted feature map $\tilde{\mathcal{X}}_{S} \in\R^{\cc\times \hh\times \ww}$, we use the following formulation,
 \begin{equation}
\tilde{\mathcal{X}}_{S} = W(\mathcal{X}_{S}) \odot \mathcal{X}_{S},
\end{equation}
where $\odot$ represents Hadamard product (i.e., element-wise product).
This operation acts like a spatial attention mechanism since the more discriminative locations related to the tracked object are weighted more heavily compared to other regions in the background. This largely improves the tracking performance as demonstrated in the experiments.
Then, we applied a global average pooling on $\tilde{\mathcal{X}}_{S}$ to obtain discriminative pooling weights $\mathcal{X}_{AWP} = g(\tilde{\mathcal{X}}_{S}) \in\R^{\cc\times 1\times 1}$ by using,
\begin{equation}
g(\tilde{\mathcal{X}}_{S})=\frac{1}{h\times w}\sum_{i=1}^h\sum_{j=1}^w\tilde{\mathcal{X}}_{S}(i,j).
\end{equation}

Finally, we apply 1D convolution and sigmoid operations to $\mathcal{X}_{AWP}$ to get our channel attention map  $\mathcal{W}_{C} \in\R^{\cc\times 1\times 1}$,
\begin{equation}
W_{C} = \sigma(Conv1D_{k}(\mathcal{X}_{AWP})),
\end{equation}
where $k=5$ is the kernel size of 1D convolution and $\sigma (.)$ is the Sigmoid function. 
At the end, the fused feature map $\mathcal{X}_{S}$ is multiplied with the channel attention map $\mathcal{W}_{C}$ to produce the aggregated feature map $\mathcal{X}_{FAM} \in\R^{\cc\times \hh\times \ww}$,
\begin{equation}
\mathcal{X}_{FAM} = \mathcal{X}_{S} \otimes  \mathcal{W}_{C}.
\end{equation}

\subsection{Implementation Details}
\smallskip\noindent{\bf Offline training.} For offline training, we use the training splits of the GOT-10k \cite{R114}, LaSOT \cite{R112} and TrackingNet \cite{R116} datasets. 
We used ResNet-18 and ResNet-50 backbones for both 2D-CNN and 3D-CNN streams. We pretrained our 3D-ResNet architecture on Kinetics dataset \cite{R108} using 4 consecutive frames to learn motion information of the moving targets. For 2D-ResNet, we initialize network weights with model pre-trained on ImageNet. Image pixel values are normalized to lie in the range [0,1]. 
During offline training, we freeze all the weights of the 3D-CNN backbone so that it does not lose its ability to generate temporal information due to the training video frames where no motion occurs. Also we freeze all weights of the 2D-CNN backbone as in \cite{R97}.

In a batch, for both train and test branches, we use 64 images and 64 4-frame clips as input to 2D-CNN and 3D-CNN backbones, respectively. Our sampling strategy can be defined as follows: we first determine the key frame to extract spatial features using 2D-CNN, and then we create a corresponding video clip with 4 consecutive frames in time order, with the key frame being the last frame. Therefore, when the frame number of key frame is smaller than 4, we apply same padding to the beginning of the clip in time dimension. We also apply data augmentations including, color jittering and image flipping. In order to ensure pixel-wise correspondence, the same data augmentations are applied to the key frame and clip. We trained FAM and IoU predictor modules for 50 epochs using Adam \cite{R118} optimizer with initial learning rates of $5\times10^{-4}$ and $10^{-3}$, respectively, and using a factor of 0.2 decay at every 15 epochs. The mean-squared error is used as loss function.


\smallskip\noindent{\bf Online tracking.} 
We employ the online tracking strategy of the ATOM tracker. During online tracking, our two-stream network is initialized with offline pre-trained model weights and all weights of our network except for the target classification module are frozen. Spatial resolution of $288\times288$ is used for input images. While the target estimation module receives layer-2 and layer-3 features of both 2D-ResNet and 3D-ResNet as input, the target classification module receives only layer-3 features of both 2D-ResNet and 3D-ResNet as input. Downsampling rates $d$ of the backbones are 8 and 16 for layer-2 and layer-3, respectively. The target classification module learns the weights of 2-layer fully convolutional network  $\wn=\left\{\wn_1,\:\wn_2\right\}$ using the $L^2(.)$ loss function as given in Eq. (\ref{eq:2}). While the first layer applies a $1\times1$ convolutional layer to reduce the channel size of layer-3 features to 64, the second layer of the classification module applies a convolutional layer with $4\times 4$ kernel size.
For the fast minimization of the objective function given in Eq. (\ref{eq:2}), we use the Gauss-Newton approach proposed in ATOM during online learning. In the initial frame, while the classification module learns both $\wn_1$ and $\wn_2$, the network updates just the weights of $\wn_2$ at every 10th frame.

We utilize data augmentation strategies such as rotation, blur, dropout, varying degrees of translation, image flipping, and color jittering to construct an initial training set including 30 samples using given the first frame. During online tracking, the new frame is determined as the key frame in time order. We use 4 frames to create a video clip with the key frame being the last frame as in offline training. Also, we copy the initial frame to complete the video clip when the key frame number is less than 4. 

\section{Experiments}

The proposed tracker is implemented by using PyTorch. On Tesla V100 GPU, TRAT runs at 21 fps and 28 fps using backbones of ResNet-18 and ResNet-50, respectively. We evaluate TRAT on eight  different benchmark datasets, including the OTB-100 \cite{wu2015object}, UAV123 \cite{R115}, NfS \cite{R113}, VOT2018 \cite{R117}, VOT2019 \cite{R92}, TrackingNet \cite{R116}, LaSOT \cite{R112} and GOT-10k \cite{R114}. 

On the OTB-100 \cite{wu2015object}, UAV123 \cite{R115} and NfS \cite{R113} datasets, we report the results in one-pass evaluation (OPE) protocol with both precision and success plots, by using OTB evaluation toolkit. 
Tracking algorithms are ranked based on the area-under-curve (AUC) scores obtained from the precision and success plots. On the VOT2018 \cite{R117} and VOT2019 \cite{R92} datasets, we use VOT challenge protocol which uses reset-based methodology in which a tracker is re-initialized whenever the tracking fails. The accuracy is measured in terms of expected average overlap (EAO), which quantitatively reflects both bounding box overlap ratio (accuracy-A) and re-initialization times (robustness-R). 
We report the AUC scores as well as precision (Pre) and normalized precision (Pre\textsubscript{norm}) scores on the TrackingNet \cite{R116} and LaSOT \cite{R112} datasets. Finally, to evaluate results on the GOT-10k \cite{R114} dataset, we calculate average overlap (AO) and success rates (SR) at overlap thresholds set to 0.5 and 0.75.

To ensure the reliability of the results, we run the proposed tracker 5 times on the OTB-100, UAV123, NfS, LaSOT, 3 times on the GOT-10k, 1 time on the TrackingNet, 15 times on both VOT2018 and VOT2019 datasets then we report the averages of these results.

\begin{table*}[t!]
\centering
\caption{Ablation study on the OTB-100, NfS and UAV123 datasets. (The red fonts indicate the best result.)}
\resizebox{0.8\textwidth}{!}{
\begin{tabular}{ cccccccccccccc }
\hline
       &        &        &        &        &        &      &     & \multicolumn{2}{c}{OTB-100}       & \multicolumn{2}{c}{NfS}           & \multicolumn{2}{c}{UAV123}         \\
Res-18 & Res-50 & 2D-CNN & 3D-CNN & Sum F. & Cat F. & Attn & AWP & AUC          & Pre                & AUC             & Pre             & AUC             & Pre              \\  \hline
\chm   &        & \chm   &        &        &        &      &     & 66.3         & 87.3               & 58.5            & 69.5            & 64.2            & 84.3             \\  \hline
       & \chm   & \chm   &        &        &        &      &     & 67.2         & 87.4               & 59.6            & 70.2            & 64.9            & 84.9             \\  \hline
\chm   &        &        & \chm   &        &        &      &     & 60.9         & 79.8               & 55.3            & 65.2            & 57.3            & 78.2             \\  \hline
       & \chm   &        & \chm   &        &        &      &     & 61.5         & 81.1               & 56.4            & 65.8            & 58.9            & 79.8             \\  \hline
       & \chm   & \chm   & \chm   & \chm   &        &      &     & 67.9         & 87.5               & 61.3            & 72.4            & 64.3            & 84.1             \\  \hline
       & \chm   & \chm   & \chm   &        & \chm   &      &     & 67.8         & 87.4               & 61.5            & 72.7            & 65.2            & 85.2             \\  \hline
       & \chm   & \chm   & \chm   &        & \chm   & \chm &     & 68.4         & 88.3               & 62.6            & 73.1            & 65.8            & 86.2             \\  \hline
       & \chm   & \chm   & \chm   &        & \chm   & \chm & \chm& \tclr{red}{68.6}& \tclr{red}{88.7}& \tclr{red}{62.8}& \tclr{red}{73.6}& \tclr{red}{66.2}& \tclr{red}{86.8} \\  \hline
\end{tabular}}
\vspace{-0.3cm}
\label{table:Ablation}
\end{table*}

\subsection{Ablation Studies}
We conduct an ablation study on the OTB-100 \cite{wu2015object}, NfS \cite{R113}, and UAV123 \cite{R115} datasets, to show the impact of individual components of the proposed tracker TRAT. Table~\ref{table:Ablation} shows the results of the used components on these datasets.

\noindent{\bf Impact of Backbone.} We take ATOM tracker as baseline to analyze the impact of using ResNet-18 and ResNet-50 as its backbone. As expected, usage of the ResNet-50 backbone improves AUC scores by about 0.7-1\% and precision scores by about 0.1-0.6\%.

We also explore the applicability of 3D-CNNs for object tracking problem. Therefore, we experiment using only 3D-CNN backbone. Although the architecture with 3D-CNN performs inferior compared to the architecture with 2D-CNN backbone, it still outperforms some of the recent trackers such as ECO, MDNet and UPDT on AUC and precision metrics for NfS dataset, where the motion of the tracked objects is significant. However, we will see in the next ablation study that 3D-CNN backbone produces complementary features to 2D-CNN features. As expected, deeper 3D-CNN backbone again achieves better performance.

\noindent{\bf Impact of Fusion Strategies.} We conduct experiments for both sum fusion and concatenation fusion in order to explore the impact of feature aggregation. As given in Table \ref{table:Ablation}, fusion of 2D-CNN and 3D-CNN features improves both AUC and precision scores on the all datasets. Concatenation fusion strategy has more improvements than sum fusion strategy on the NfS and UAV123 datasets. On the OTB-100 dataset, both fusion approaches obtain similar results. Therefore, we have used concatenation fusion for the rest of the ablation studies.

\noindent{\bf Impact of Chanel Attention.} After aggregating the feature maps using concatenation fusion, we have applied the channel attention module to weight the importance of aggregated channels. For this analysis, we have used global average pooling instead of AWP. Channel attention module improves AUC scores by about 0.6-1.0\% and precision scores by about 0.4-1.0\%.

\noindent{\bf Impact of AWP.} Finally, we have investigated the impact of the AWP component. It improves AUC scores by about 0.2-0.4\% and precision scores by around 0.4-0.6\%. To compare with other state-of-the-art trackers, we utilize this final tracker, which is called TRAT.

\subsection{State-of-the-art Comparison}
\smallskip\noindent{\bf Object Tracking Benchmark (OTB-100).} The OTB-100 \cite{wu2015object} is a well-known single object tracking dataset.  This dataset includes 100 fully annotated video sequences, with 11 various challenging factors. We report results in Table \ref{table:State} using the OPE protocol. UPDT \cite{R119} and SiamRPN++ \cite{R88} trackers achieve the best accuracy and precision scores, respectively. Our proposed method obtains competitive results on this dataset.

\smallskip\noindent{\bf NfS.} We evaluate our approach on need for speed dataset \cite{R113} (30 FPS version), which is the first high frame rate dataset recorded at real-world scenes. It includes 100 fully annotated videos (380K frames) with fast moving target objects. The AUC and precision scores of state-of-the-art trackers are shown in Table \ref{table:State}. TRAT achieves a success score of 62.8\% which is higher than the previous best result. In terms of precision, TRAT obtains the best results after DiMP50 \cite{R100} tracker. Moreover, TRAT outperforms our baseline tracker, ATOM \cite{R97}, with relative gains of 7.4\% and 5.9\% in terms of AUC and precision scores, respectively. These results demonstrate the importance of using 3D-CNN in videos involving a fast-moving target object.

\begin{figure}[tb]
	\centering
	\includegraphics[scale=0.115]{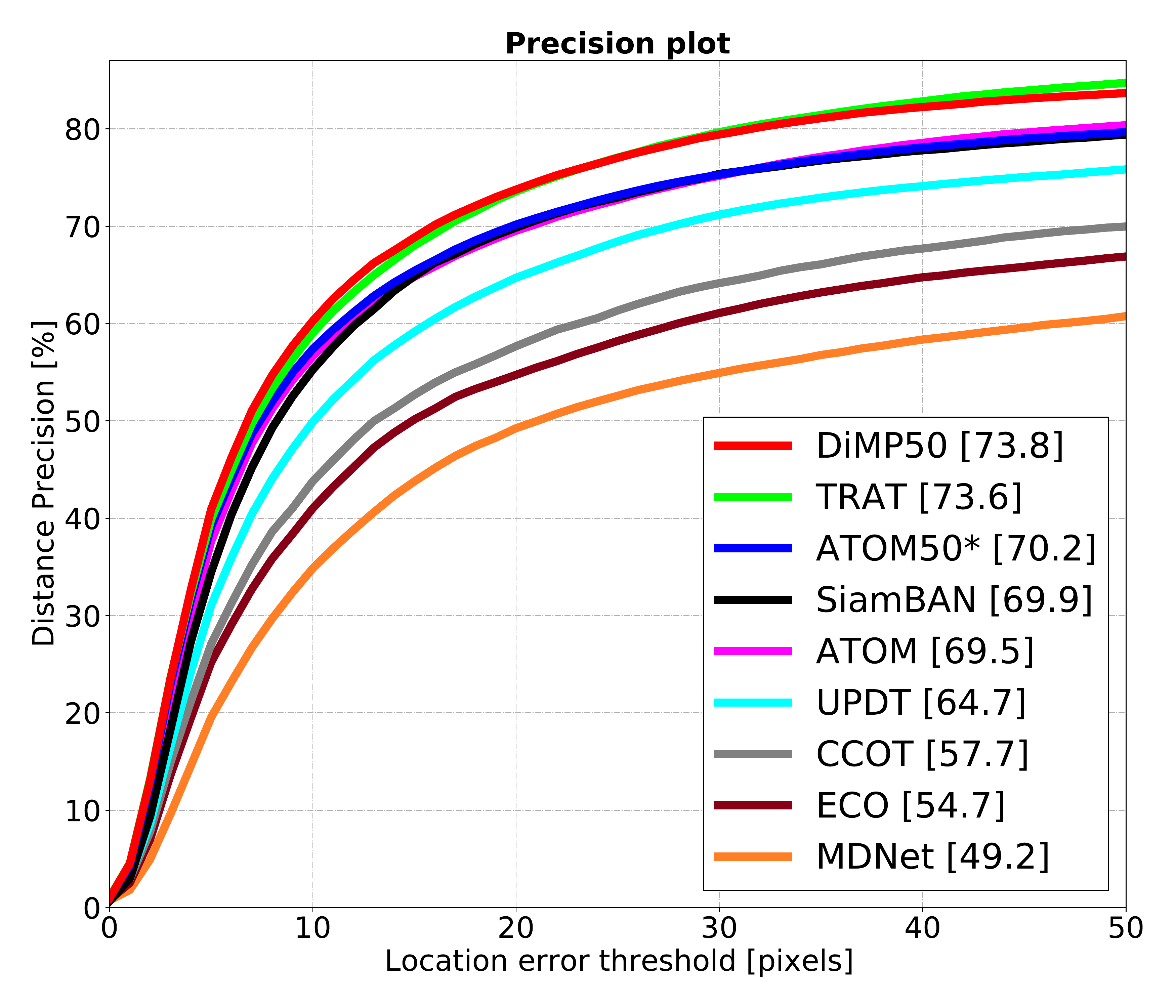}
	\includegraphics[scale=0.115]{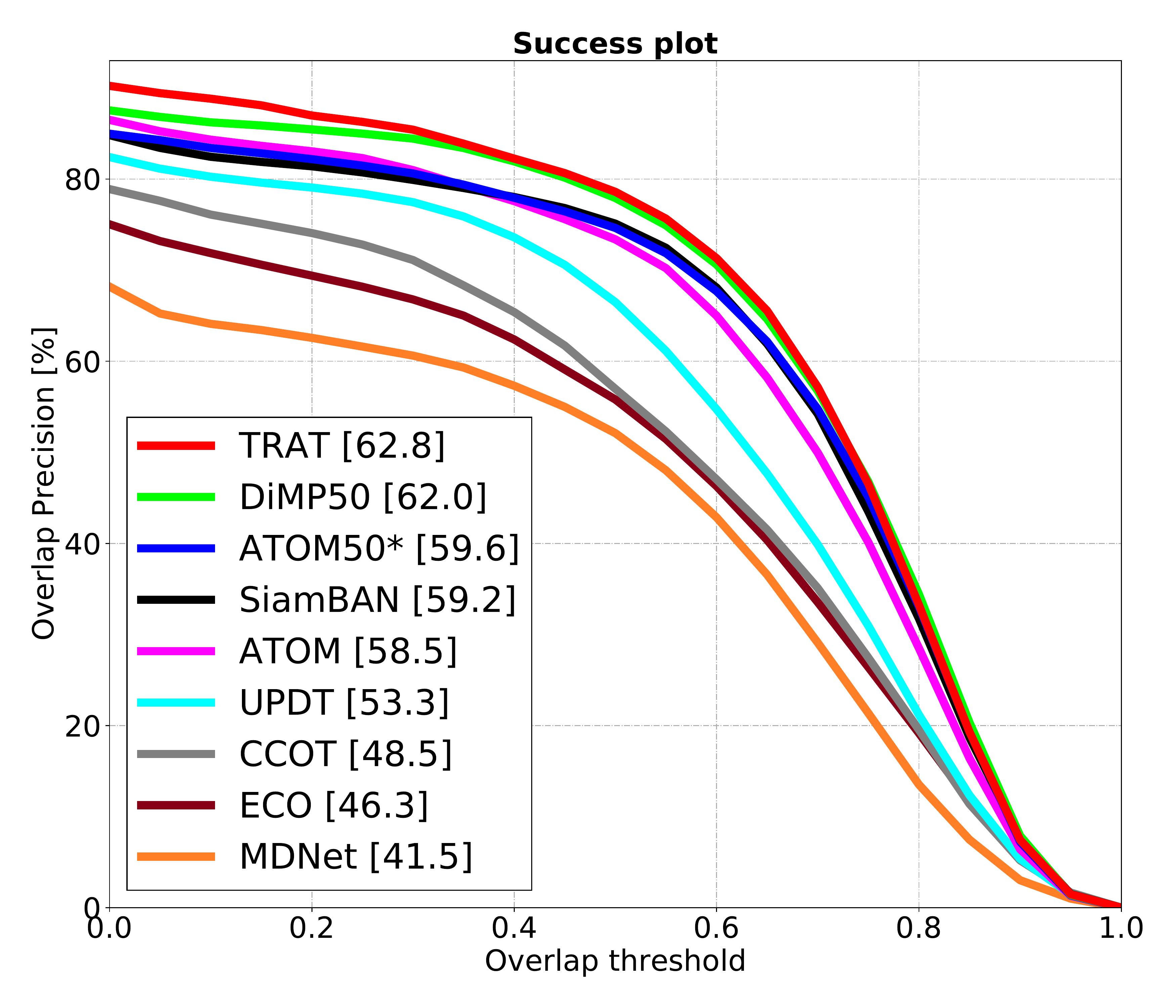}
	\caption{Precision and success plots of the TRAT and state-of-the-art trackers on the NfS using one pass evaluation (OPE) protocol.}
	\vspace{-0.3cm}
	\label{fig:NfS}
\end{figure}

\smallskip\noindent{\bf UAV123.} The UAV123 \cite{R115} includes 123 videos captured from low-altitude unmanned aerial vehicles. We report the success and precision scores in Table \ref{table:State}. Our tracker achieves an AUC score of 66.2\% and precision score of 86.2\%, which are higher than the previous best performing trackers DiMP50 \cite{R100} with 65.3\% and 86.0\% in terms of AUC and precision scores, respectively. TRAT outperforms ATOM \cite{R97} considerably by AUC score of 63.1\% $\rightarrow$ 66.2\% and precision score of 84.1\% $\rightarrow$ 86.2\%.

\begin{table}[H]
\centering
\caption{State-of-the-art comparison on the OTB-100, NfS and UAV123 datasets. (The red and blue fonts respectively indicate the best and the second best results.)}
\resizebox{\columnwidth}{!}{\begin{tabular}{ lccccccc }\hline
                        & \multicolumn{2}{c}{OTB-100}          & \multicolumn{2}{c}{NfS}              & \multicolumn{2}{c}{UAV123}          \\
                        & AUC              & Precision         & AUC               & Precision        & AUC              & Precision        \\  \hline
 UPDT \cite{R119}       & \tclr{red}{70.7} & -                 & 54.1              & -                & 55.0             & -                \\  \hline
 RankingT \cite{R103}   & \tclr{blue}{70.1}& \tclr{blue}{91.1} & -                 & -                & 53.5             & 74.6             \\  \hline
 DaSiam-RPN \cite{R120} & 65.8             & 88.0              & 60.6              & 75.2             & 56.9             & 78.1             \\  \hline
 MDNet \cite{R17}       & 67.8             & 90.9              & 41.5              & 49.2             & 56.9             & 78.1             \\  \hline
 ECO \cite{R27}         & 69.1             & 91.0              & 46.6              & -                & 52.5             & 74.1             \\  \hline
 SiamRPN++ \cite{R88}   & 69.6             & \tclr{red}{91.4}  & -                 & -                & 61.3             & 80.7             \\  \hline
 SiamBAN \cite{R121}    & 69.6             & 91.0              & 59.4              & -                & 63.1             & 83.3             \\  \hline
 DiMP50 \cite{R100}     & 68.4             & 89.4              & \tclr{blue}{62.0} & \tclr{red}{73.7}& \tclr{blue}{65.3}& \tclr{blue}{86.0}\\  \hline
 ATOM \cite{R97}        & 66.3             & 87.4              & 58.5              & 69.5             & 63.1             & 84.3             \\  \hline
 TRAT                   & 68.6             & 88.7              & \tclr{red}{62.8}  & \tclr{blue}{73.6} & \tclr{red}{66.2} & \tclr{red}{86.8} \\  \hline
\end{tabular}}
\vspace{-0.3cm}
\label{table:State}
\end{table}

\smallskip\noindent{\bf TrackingNet.} We evaluate our tracker on the large-scale TrackingNet \cite{R116} benchmark using success, precision and normalized precision metrics. Table \ref{table:TrackingNet} shows the results of state-of-the-art trackers on the TrackingNet test split which contains 511 videos. TRAT ranks second for all metrics and achieves competitive results compared to the best performing trackers, SiamRPN++ \cite{R88} and SiamAttn \cite{R123}.

\begin{table}[H]
\centering
\caption{State-of-the-art comparison on the TrackingNet dataset. (The red and blue fonts respectively indicate the best and the second best results.)}
\resizebox{\columnwidth}{!}{\begin{tabular}{ lcccccccc }
\hline
                              & UPDT \cite{R119}& DaSiam-RPN \cite{R120}& D3S \cite{R122}& SiamAttn \cite{R123}& SiamRPN++ \cite{R88}& DiMP50 \cite{R100}& ATOM \cite{R97}& TRAT \\  \hline
 Pre (\%)                     & 55.7            & 59.1                  & 66.4           & -                   & \tclr{red}{69.4}    & 68.7              & 64.8           & \tclr{blue}{68.9}\\  \hline
 Pre\textsubscript{norm} (\%) & 70.2            & 73.3                  & 76.8           & \tclr{red}{81.7}    & 80.0                & 80.1              & 77.1           & \tclr{blue}{80.3}\\  \hline
 Success (\%)                 & 61.1            & 63.8                  & 72.8           & \tclr{red}{75.2}    & 73.3                & 74.0              & 70.3           & \tclr{blue}{74.2}\\  \hline
\end{tabular}}
\vspace{-0.3cm}
\label{table:TrackingNet}
\end{table}

\smallskip\noindent{\bf LaSOT.} Large scale object tracking \cite{R112} dataset consist of 1400 sequences with 3.52 million frames, which have high-quality dense annotations. We evaluate our tracker on test split which consist of 280 videos. TRAT achieves the second best result on success metric and competitive results for normalized precision metric. More importantly, TRAT significantly outperforms ATOM by 7.0\% and 5.1\% on normalized precision and success metrics, respectively.

\begin{figure}[tb]
	\centering
	\includegraphics[scale=0.115]{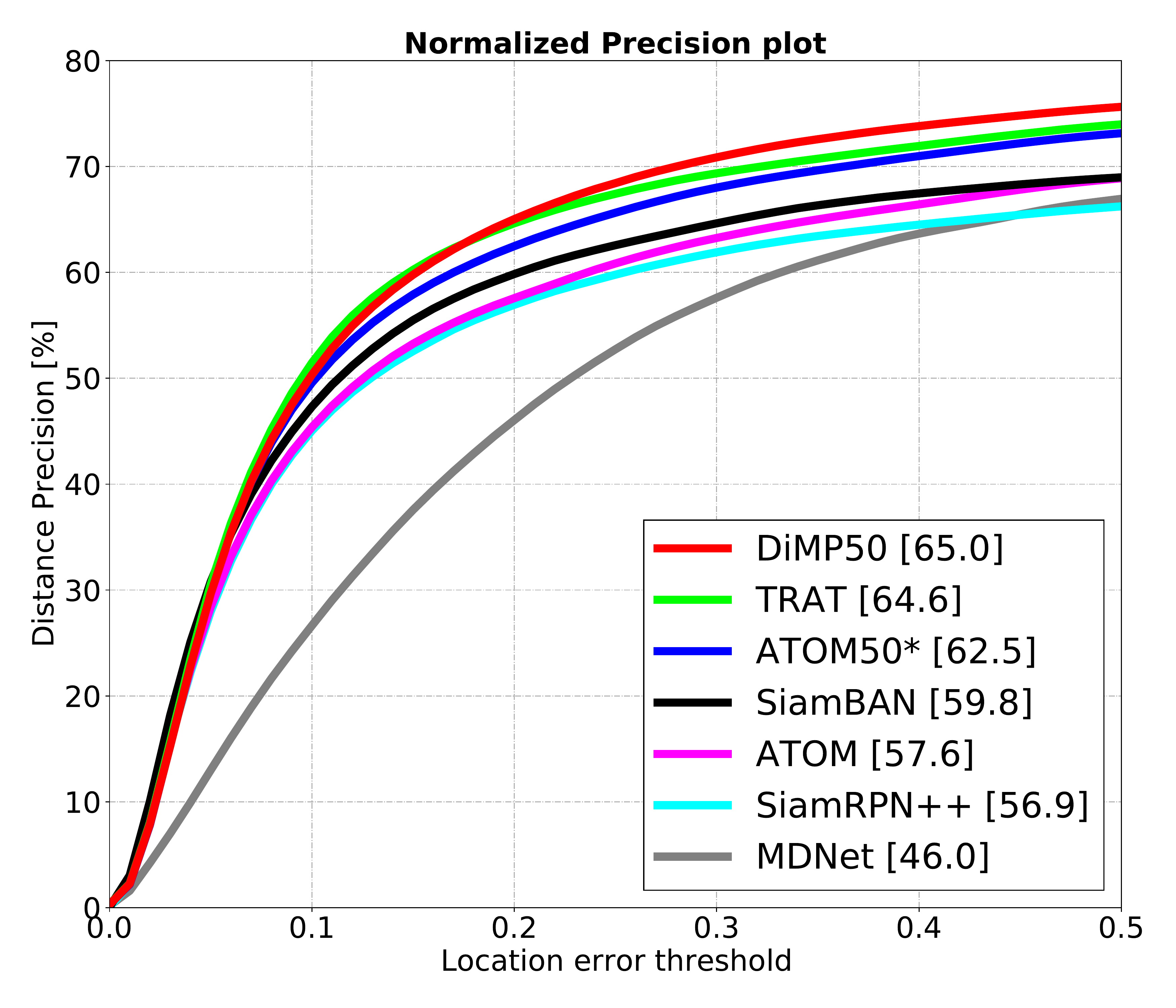}
	\includegraphics[scale=0.115]{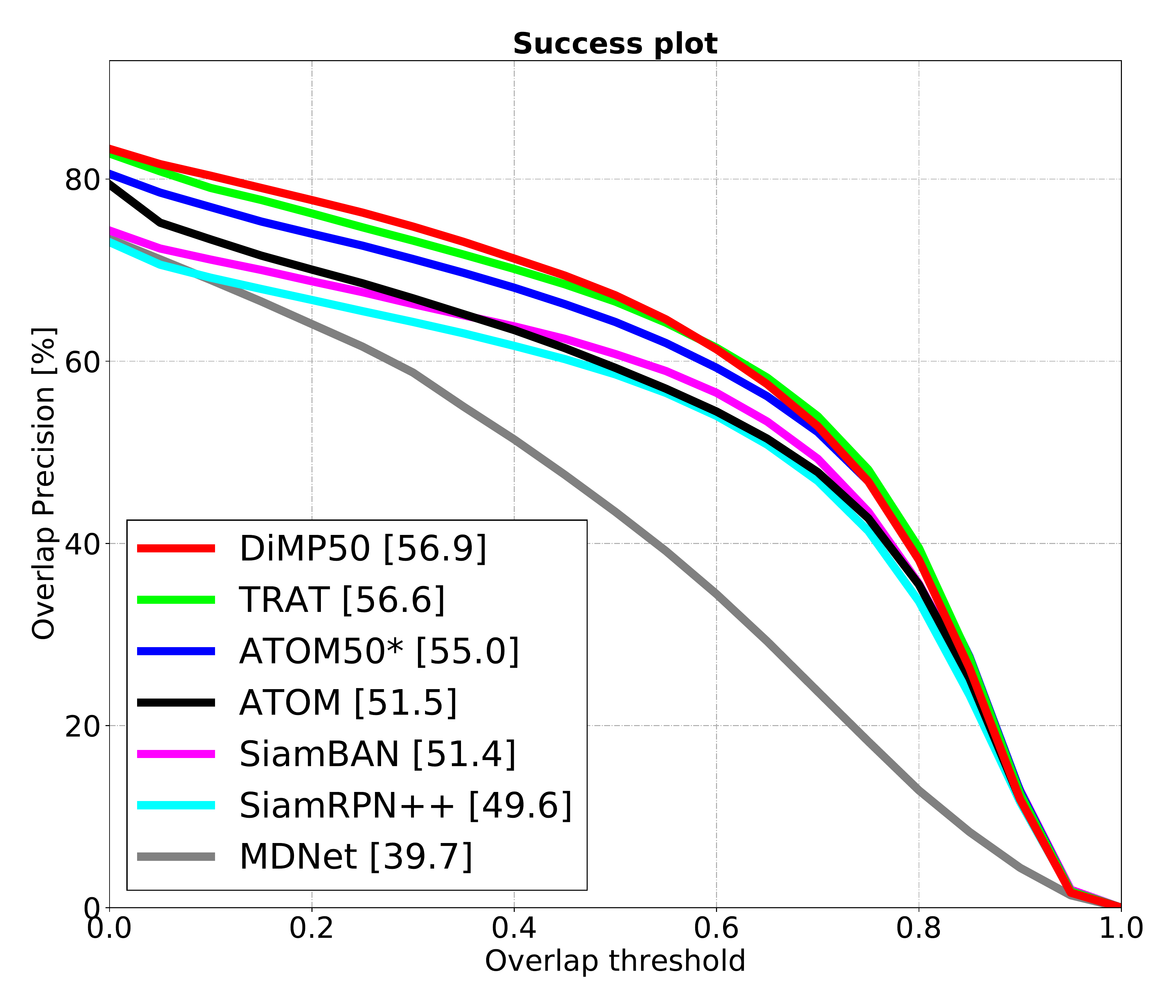}
	\caption{Normalized precision and success plots of the TRAT and state-of-the-art trackers on the LaSOT.}
	\label{fig:LaSOT}
	\vspace{-0.3cm}
\end{figure}

\begin{table}[H]
\centering
\caption{State-of-the-art comparison on the LaSOT dataset. (The red and blue fonts respectively indicate the best and the second best results.)}
\resizebox{\columnwidth}{!}{\begin{tabular}{ lcccccccc }
\hline
                             & DaSiam-RPN \cite{R120}& SiamBAN \cite{R121}& SiamAttn \cite{R123}& SiamRPN++ \cite{R88}& DiMP50 \cite{R100}& ATOM \cite{R97}& TRAT \\  \hline
Pre\textsubscript{norm} (\%) & 49.6                  & 59.8               & \tclr{blue}{64.8}                & 56.9                & \tclr{red}{64.8}              & 57.6           & 64.6 \\  \hline
Success (\%)                 & 41.5                  & 51.4               & 56.0                & 49.6                & \tclr{red}{56.8}              & 51.5           & \tclr{blue}{56.6} \\  \hline
\end{tabular}}
\vspace{-0.3cm}
\label{table:LaSOT}
\end{table}

\smallskip\noindent{\bf GOT-10k.} The GOT-10k dataset \cite{R114} is another large-scale object tracking dataset which contains 10K video sequences for the training set and 180 videos with an average length of 127 frames for the test set. We evaluate TRAT on the GOT-10k dataset using the GOT-10k evaluation toolkit. Table \ref{table:Got10k} shows the state-of-the-art comparison on the GOT-10k test split. We report the results in terms of success rates (SR) and average overlap (AO).
In terms of SR at overlap thresholds set to 0.5 and 0.75, TRAT obtains the best and second-best results of 72.0\% and 46.7\% respectively. Again TRAT has a comparable AO result to DiMP50 \cite{R100}. Compared with the baseline tracker ATOM \cite{R97}, TRAT has significant improvements of 8.6\% on SR$_{0.50}$, 6.5\% on SR$_{0.75}$, and 5.2\% on AO.

\begin{table}[H]
\centering
\caption{State-of-the-art comparison on the GOT-10k dataset. (The red and blue fonts respectively indicate the best and the second best results.)}
\resizebox{\columnwidth}{!}{\begin{tabular}{ lcccccccc }
\hline
                  & MDNet \cite{R17}& CCOT \cite{R26}& ECO \cite{R27}& SiamMask \cite{R91}& D3S \cite{R122}& DiMP50 \cite{R100}& ATOM \cite{R97}& TRAT              \\  \hline
 SR$_{0.50}$ (\%) & 30.3            & 32.8           & 30.8          & 58.7               & 67.6           & \tclr{blue}{71.7} & 63.4           & \tclr{red}{72.0}  \\   \hline
 SR$_{0.75}$ (\%) & 9.9             & 10.7           & 11.1          & 36.6               & 46.2           & \tclr{red}{49.2}  & 40.2           & \tclr{blue}{46.7} \\   \hline
 AO (\%)          & 29.9            & 32.5           & 31.6          & 51.4               & 59.7           & \tclr{red}{61.1}  & 55.6           & \tclr{blue}{60.8} \\  
\hline
\end{tabular}}
\vspace{-0.3cm}
\label{table:Got10k}
\end{table}

\smallskip\noindent{\bf VOT2018-VOT2019.} We evaluate trackers on the short-term challenges of the VOT2018 \cite{R117} and VOT 2019 \cite{R92} datasets which consist of 60 fully annotated video sequences by rotated bounding boxes.  All frames are also annotated for different challenging factors including motion change, size change, occlusion, illumination change, camera motion, and unassigned which did not correspond to the other attributes. We run our trackers 15 times to extract results by provided VOT evaluation toolkit. Table~\ref{table:VOT-2018} and Table~\ref{table:VOT-2019} show the state-of-the-art comparison on the VOT2018 and VOT2019 datasets, respectively. We report all results using a reset-based methodology which contains accuracy (A), robustness (R) and expected average overlap (EAO) metrics for both datasets. Our proposed method TRAT outperforms the state-of-the-art trackers with 0.456 EAO, 0.605 accuracy, and 0.148 robustness scores on the VOT2018 dataset. Especially in terms of EAO, TRAT achieves large gains of 5.5\%, compared to our baseline method ATOM \cite{R97}.

\begin{table}[H]
\centering
\caption{State-of-the-art comparison on the VOT2018 dataset.}
\resizebox{\columnwidth}{!}{\begin{tabular}{ lccccccc }
\hline
            & RankingT \cite{R103}& SiamBAN \cite{R121}& DaSiam-RPN \cite{R119}& SiamRPN++ \cite{R88}& DiMP50 \cite{R100}& ATOM \cite{R97}& TRAT              \\  \hline
 AUC        & 0.554               & 0.597              & 0.503                 &\tclr{blue}{0.604}   & 0.597             & 0.590          & \tclr{red}{0.605} \\  \hline
 Robustness & 0.213               & 0.178              & 0.159                 & 0.234               &\tclr{blue}{0.153} & 0.204          & \tclr{red}{0.148} \\  \hline
 EAO        & 0.335               &\tclr{blue}{0.452}  & 0.389                 & 0.414               & 0.440             & 0.401          & \tclr{red}{0.456} \\  \hline
\end{tabular}}
\vspace{-0.3cm}
\label{table:VOT-2018}
\end{table}

On the VOT2019 dataset, TRAT achieves the best accuracy score of 0.606. Moreover, in terms of robustness and EAO, it obtains the second-best scores of 0.302 and 0.381, respectively.

\begin{table}[H]
\centering
\caption{State-of-the-art comparison on the VOT2019 dataset.}
\resizebox{\columnwidth}{!}{\begin{tabular}{ lccccccc }
\hline
            & RankingT \cite{R103}& SiamBAN \cite{R121}& SiamMask \cite{R91}& SiamRPN++ \cite{R88}& DiMP50 \cite{R100}& ATOM \cite{R97}& TRAT          \\  \hline
 AUC        & 0.528                & \tclr{blue}{0.602} & 0.594              & 0.599               & 0.594             & 0.603          & \tclr{red}{0.606} \\  \hline
 Robustness & 0.360                & 0.327              & 0.461              & 0.482               & \tclr{red}{0.278} & 0.411          & \tclr{blue}{0.302} \\  \hline
 EAO        & 0.270                & \tclr{red}{0.396}  & 0.287              & 0.285               & 0.379             & 0.292          & \tclr{blue}{0.381} \\  \hline
\end{tabular}}
\vspace{-0.3cm}
\label{table:VOT-2019}
\end{table}

\section{Conclusion}
This paper introduces a two-stream deep neural network tracker using 2D-CNN and 3D-CNN backbones. 
The proposed tracker is built based on ATOM tracker architecture, and the 2D-CNN stream is used to capture appearance information whereas the 3D-CNN is used for capturing motion cues.
The feature maps coming from these two streams are fused by using an effective channel fusion and attention mechanism that considers only local correlations between feature channels. This significantly decreases the computational complexity, and allows the network to learn the feature aggregation parameters during online tracking efficiently.
The proposed tracker, TRAT, is evaluated on most of the tracking datasets and results show that it achieves the state-of-the-art accuracies on most of the tested datasets. Moreover, it significantly  outperforms the baseline tracker ATOM that uses only appearance information alone. This indicates the importance of motion information in tracking applications. 

{\small
\bibliographystyle{ieee_fullname}
\bibliography{reff}
}

\end{document}